\documentclass[10pt,twocolumn,letterpaper]{article}

\usepackage{cvpr}
\usepackage{times}
\usepackage{epsfig}
\usepackage{graphicx}
\usepackage{amsmath}
\usepackage{amssymb}
 \usepackage[table,xcdraw]{xcolor}
\usepackage{multirow}
\usepackage{siunitx,booktabs,etoolbox}
\usepackage{tabularx}
\usepackage{ragged2e}
\usepackage{bbm}
\usepackage{glossaries}
\usepackage{enumitem}
\usepackage{placeins}
\usepackage{bm}

\usepackage[tableposition=above]{caption}

\usepackage[pagebackref=true,breaklinks=true,letterpaper=true,colorlinks,bookmarks=false]{hyperref}

\cvprfinalcopy 

\def\cvprPaperID{4919} 

\newcommand{\tbl}{Table}

\newcommand{\qcg}{QCG}
\newglossaryentry{VQAv1}
{
    name=VQAv1,
    description={VQAv1}
}
\newglossaryentry{VQAv2}
{
    name=VQAv2,
    description={VQAv2}
}
\newglossaryentry{CVQA}
{
    name=CVQA,
    description={Compositional VQA}
}
\newglossaryentry{VQACP}
{
    name=VQACPv2,
    description={VQA under Changing Priors}
}
\newglossaryentry{VQACPv2}
{
    name=VQACPv2,
    description={VQA under Changing Priors}
}
\newglossaryentry{TDIUC}
{
    name=TDIUC,
    description={Task Driven Image Understanding Challenge}
}
\newglossaryentry{CLEVR}
{
    name=CLEVR,
    description={Compositional Language and Elementary Visual Reasoning}
}
\newglossaryentry{CLEVR-Humans}
{
    name=CLEVR-Humans,
    description={Compositional Language and Elementary Visual Reasoning}
}
\newglossaryentry{CoGenT}
{
    name=CLEVR-CoGenT,
    description={CLEVR Generalization Tests}
}
\newglossaryentry{UpDn}
{
    name=UpDn,
    description={Bottom Up Attention Model}
}
\newglossaryentry{QCG}
{
    name=QCG,
    description={Question Conditioned Graph}
}
\newglossaryentry{BAN}
{
    name=BAN,
    description={Bilinear Attention Network}
}
\newglossaryentry{MAC}
{
    name=MAC,
    description={Memory, Attention and Control}
}
\newglossaryentry{RN}
{
    name=RN,
    description={Relation Network}
}
\newglossaryentry{RAMEN}
{
    name=RAMEN,
    description={Recurrent Aggregation of Multimodal Embeddings Network}
}

\begin{document}

\title{Answer Them All! Toward Universal Visual Question Answering Models}

\author{Robik Shrestha$^1$\qquad Kushal Kafle$^1$\qquad Christopher Kanan$^{1,2,3}$\\ 
$^1$Rochester Institute of Technology\qquad $^2$PAIGE \qquad $^3$Cornell Tech\\
{\tt\small \{rss9369, kk6055,  kanan\}@rit.edu}
}

\maketitle

\begin{abstract}
   Visual Question Answering (VQA) research is  split into two camps: the first focuses on VQA datasets that require natural image understanding and the second focuses on synthetic datasets that test reasoning. A good VQA algorithm should be capable of both, but only a few VQA algorithms are tested in this manner. We compare five state-of-the-art VQA algorithms across eight VQA datasets covering both domains. To make the comparison fair, all of the models are standardized as much as possible, \eg, they use the same visual features, answer vocabularies, etc. We find that methods do not generalize across the two domains. To address this problem, we propose a new VQA algorithm that rivals or exceeds the state-of-the-art for both domains.
\end{abstract}

\begin{figure}[h]
    \centering
    \includegraphics[width=0.46\textwidth]{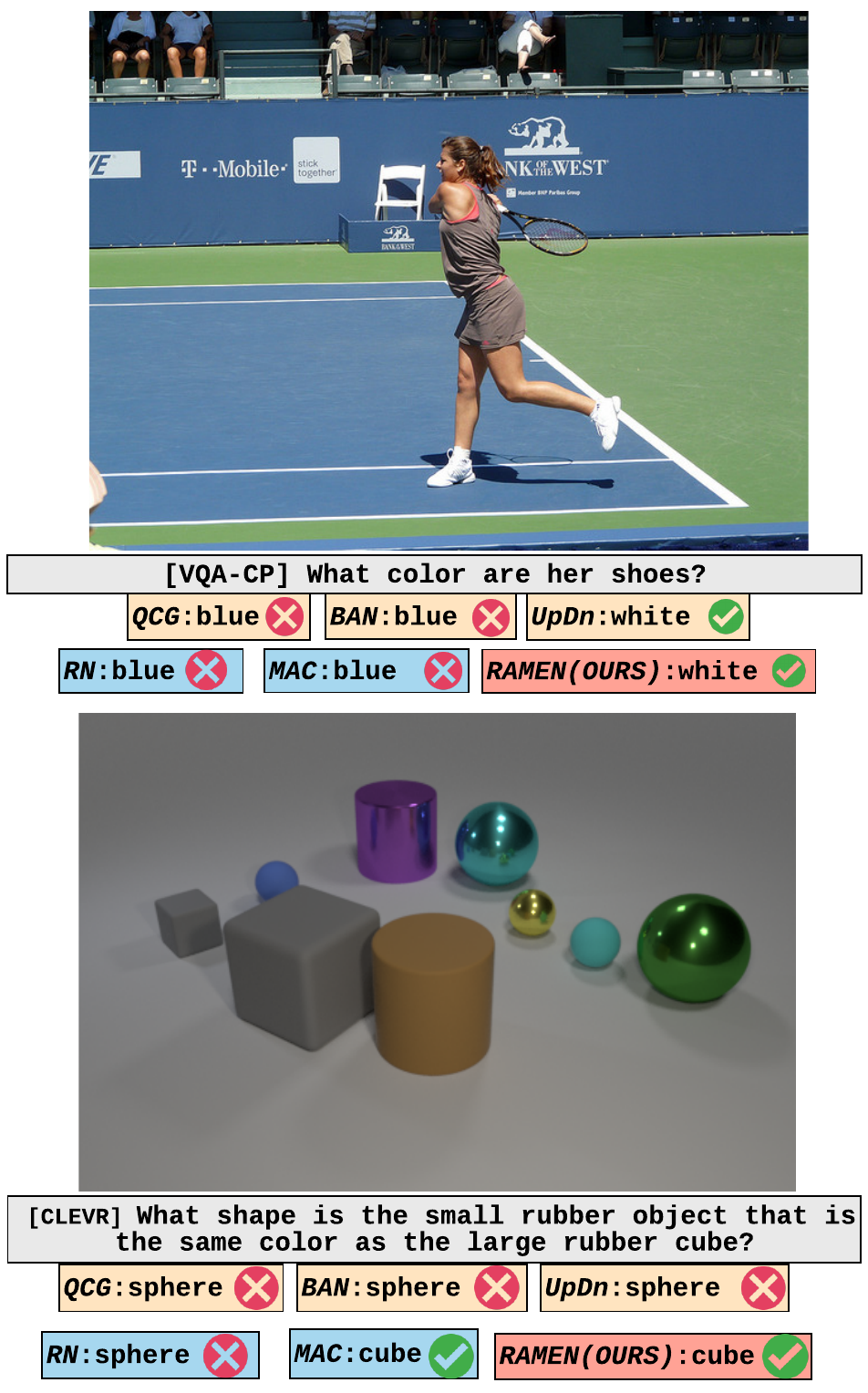}
    \caption{Many VQA algorithms do not transfer well across natural and synthetic datasets. We argue it is necessary to do well on both domains and present an algorithm that achieves this goal.}
    \label{fig:vqa2018_main}
\end{figure}

\section{Introduction}

Visual Question Answering (VQA) requires a model to understand and reason about visuo-linguistic concepts to answer open-ended questions about images. Correctly answering these questions demands numerous capabilities, including object localization, attribute detection, activity classification, scene understanding,  reasoning, counting, and more.  The first VQA datasets contained real-world images with crowdsourced questions and answers~\cite{malinowski2014multi,antol2015vqa}. It was assumed that this would be an extremely difficult problem and was proposed as a form of Visual Turing Test to benchmark performance in computer vision. However, it became clear that many high performing algorithms were simply exploiting biases and superficial correlations, without really understanding the visual content~\cite{kafle2017analysis,AgrawalBP16}. For example,  answering `yes' to all yes/no questions in \gls{VQAv1}~\cite{antol2015vqa} results in an accuracy of 71\% on these questions~\cite{kafle2016review}. Later natural image VQA datasets endeavored to address this issue. By associating each question with complementary images and different answers, \gls{VQAv2}~\cite{goyal2017making} reduces some forms of language bias. \gls{TDIUC}~\cite{kafle2017analysis} analyzes generalization to multiple kinds of questions and rarer answers. \gls{CVQA}~\cite{Agrawal2017CVQAAC} tests concept compositionality and \gls{VQACP}~\cite{vqacp} tests performance when train and test distributions differ.

While later natural image datasets have reduced bias, the vast majority of questions in these datasets do not rigorously test reasoning skills. Several synthetic datasets~\cite{johnson2016clevr,AndreasRDK15} were created as a remedy. They contain simple visual scenes with challenging questions that test multi-step reasoning, counting, and logical inference. To properly evaluate an algorithm's robustness, the creators of these datasets have argued algorithms should be tested on both domains~\cite{johnson2016clevr,AndreasRDK15}. 

However, almost all recent papers report their performance on only one of these two domains. The best algorithms for CLEVR are not tested on natural image VQA datasets~\cite{hudson2018compositional,clevr-iep,Mascharka_2018_CVPR,perez2018film,nsvqa}, and vice versa~\cite{ben2017mutan,anderson2018bottom,kim2018bilinear,Nguyen_2018_CVPR,Farazi2018ReciprocalAF}. Here, we test five state-of-the-art VQA systems across eight datasets.  We found that most methods do not perform well on both domains (Fig.~\ref{fig:vqa2018_main}), with some suffering drastic losses in performance.  We propose a new model that rivals state-of-the-art methods on all of the evaluated datasets.

\paragraph{Our major contributions are:} 
\begin{enumerate}[noitemsep, nolistsep]
    \item We perform a rigorous comparison of five  state-of-the-art algorithms across eight VQA datasets, and we find that many do not generalize across domains. 
    \item Often VQA algorithms use different visual features and answer vocabularies, making it difficult to assess performance gains. We endeavor to standardize the components used across models, \eg, all of the algorithms we compare use \emph{identical} visual features, which required elevating the methods for synthetic scenes to use region proposals.
    \item We find that most VQA algorithms are not capable of understanding real-word images \emph{and} performing compositional reasoning. All of them fare poorly on generalization tests, indicating that these methods are still exploiting dataset biases.
    \item We describe a new VQA algorithm that rivals state-of-the-art methods on all datasets and performs best overall.
\end{enumerate}

\section{Related Work}

\subsection{VQA Datasets}

\begin{table}[t]
\small
\centering
\caption{Comparison of datasets used in this paper. \label{tbl:datasets-comparison}}
\begin{tabular}{@{}lrrcc@{}}
\toprule
\multicolumn{1}{l}{\textbf{Dataset}} & \textbf{\begin{tabular}[r]{@{}r@{}}Num. of \\ Images\end{tabular}} & \textbf{\begin{tabular}[r]{@{}r@{}}Num. of \\ QA Pairs\end{tabular}} & \textbf{\begin{tabular}[c]{@{}c@{}}Question\\ Source\end{tabular}} & \textbf{\begin{tabular}[c]{@{}c@{}}Image \\ Source\end{tabular}} \\ \midrule
\textbf{\gls{VQAv1}}                     & 204K                                                              & 614K                                                                & Human                                                            & Natural                                                             \\
\textbf{\gls{VQAv2}}                     & 204K                                                              & 1.1M                                                              & Human                                           & Natural                                                             \\
\textbf{TDIUC}                       & 167K                                                              & 1,6M                                                              & Both                                                                          & Natural                                              \\
\textbf{C-VQA}                       & 123K                                                              & 369K                                                                & Human                                                            & Natural                                                             \\
\textbf{\gls{VQACP}}                 & 219K                                                              & 603K                                                                & Human                                                            & Natural                                                             \\ \midrule
\textbf{CLEVR}                       & 100K                                                              & 999K                                                                & Synthetic                                                                           & Synthetic                                                        \\
\textbf{CLEVR-H}                & 32K                                                                & 32K                                                       & Human                                                            & Synthetic                                                        \\
\textbf{CoGenT-A}                    & 100K                                                              & 999K                                                                & Synthetic                                                                           & Synthetic                                                        \\
\textbf{CoGenT-B}                    & 30K                                                               & 299K                                                                & Synthetic                                                                           & Synthetic                                                        \\ \bottomrule
\end{tabular}
\end{table}

Many VQA datasets have been proposed over the past four years. Here, we briefly review the datasets used in our experiments. Statistics for these datasets are given in Table~\ref{tbl:datasets-comparison}. See \cite{kafle2016review} and \cite{wu2016visual}  for reviews.

\textbf{\gls{VQAv1}/\gls{VQAv2}.} \gls{VQAv1} \cite{antol2015vqa} is one of the earliest, open-ended VQA datasets collected from human annotators. \gls{VQAv1} has multiple kinds of language bias, including some questions being heavily correlated with specific answers.  \gls{VQAv2}~\cite{goyal2017making} endeavors to mitigate this kind of language bias by collecting complementary images per question that result in different answers, but other kinds of language bias are still present, \eg, reasoning questions are  rare compared to detection questions. Both datasets have been widely used and \gls{VQAv2} is the de facto benchmark for natural image VQA.

\textbf{\gls{TDIUC}}~\cite{kafle2017analysis} attempts to address the bias in the \emph{kinds} of questions posed by annotators by categorizing questions into 12 distinct types, enabling nuanced task-driven evaluation. It has metrics to evaluate generalization across question types.

\textbf{\gls{CVQA}}~\cite{Agrawal2017CVQAAC} is a re-split of \gls{VQAv1} to test generalization to  concept compositions not seen during training, \eg, if the train set asks about `green plate' and `red light,' the test set will ask about `red plate' and `green light.' \gls{CVQA} tests the ability to combine previously seen concepts in unseen ways.

\textbf{\gls{VQACPv2}}~\cite{vqacp} re-organizes \gls{VQAv2} such that answers for each question type are distributed differently in the train and test sets,  \eg, `blue' and `white' might be the most frequent answers to `What color...' questions in the train set, but these answers will rarely occur in the test set.  Since it has different biases in the train and test sets, doing well on \gls{VQACPv2} suggests that the system is generalizing by overcoming the biases in the training set. 

\textbf{\gls{CLEVR}}~\cite{johnson2016clevr} is a synthetically generated dataset, consisting of visual scenes with simple geometric shapes, designed to test `compositional language and elementary visual reasoning.' CLEVR's questions often require long chains of complex reasoning. To enable fine-grained evaluation of reasoning abilities, CLEVR's questions are categorized into five tasks: `querying attribute,' `comparing attributes,' `existence,' `counting,' and `integer comparison.' 
Because all of the questions are programmatically generated, the \textbf{\gls{CLEVR-Humans}}~\cite{clevr-iep} dataset was created to provide human-generated questions for CLEVR scenes to test generalization to free-form questions.

\textbf{\gls{CoGenT}} tests the ability to handle unseen concept composition and remember old concept combinations. It has two splits: CoGenT-A and CoGenT-B, with mutually exclusive shape+color combinations. If models trained on CoGenT-A perform well on CoGenT-B without fine-tuning, it indicates generalization to novel compositions. If models fine-tuned on CoGenT-B still perform well on CoGenT-A, it indicates the ability to remember old concept combinations. The questions in these datasets are more complex  than most in CVQA.

Using VQAv1 and VQAv2 alone makes it difficult to gauge whether an algorithm is capable of performing robust  compositional reasoning or whether it is using superficial correlations to predict an answer. In part, this is due to the limitations of seeking crowdsourced questions and answers, with humans biased towards asking certain kinds of questions more often for certain images, \eg, counting questions are most often asked if there are two things of the same type in a scene and almost never have an answer of zero.  While CVQA and VQACPv2 try to overcome these issues, synthetic datasets~\cite{johnson2016clevr,AndreasRDK15,kafle2018dvqa} minimize such biases to a greater extent, and serve as an important litmus-test to measure \textit{specific} reasoning skills, but the synthetic visual scenes lack complexity and variation.

Natural and synthetic datasets serve complementary purposes, and the creators of synthetic datasets have argued that both should be used, \eg, the creators of SHAPES, an early VQA dataset similar to CLEVR, wrote `While success on this dataset is by no means a sufficient condition for robust visual QA, we believe it is a necessary one'~\cite{AndreasRDK15}. While this advice has largely been ignored by the community, we \textbf{strongly believe}  it is necessary to show that VQA algorithms are capable of tackling VQA in both natural and synthetic domains with little modification. Otherwise, an algorithm's ability to generalize will not be fully assessed.

\subsection{VQA Algorithms}

Many algorithms for natural image VQA have been proposed, including Bayesian approaches~\cite{kafle2016,malinowski2014multi}, methods using spatial attention~\cite{Yang2016,LuYBP16,noh2016training,anderson2018bottom}, compositional approaches~\cite{AndreasRDK15,andreas2016learning,hu2017learning},  bilinear pooling schemes~\cite{kim2016hadamard,FukuiPYRDR16}, and others~\cite{teney2017visual,norcliffe2018learning,kafle2017data}. Spatial attention mechanisms~\cite{anderson2018bottom,LuYBP16,nam2016dual,FukuiPYRDR16,ben2017mutan} are one of the most widely used methods for natural language VQA. Attention computes relevance scores over visual and textual features allowing  models to process only relevant information.  Among these, we evaluate UpDn~\cite{anderson2018bottom}, QCG~\cite{norcliffe2018learning}, and BAN~\cite{kim2018bilinear}. We describe these algorithms in more detail in Sec.~\ref{sec:vqa-models}. 

Similarly, many methods have been created for synthetic VQA datasets. Often, these algorithms place a much greater emphasis on learning compositionality, relational reasoning, and interpretability compared to algorithms for natural images. Common approaches include modular networks, with some using ground-truth programs~\cite{clevr-iep,Mascharka_2018_CVPR}, and others learning compositional rules implicitly~\cite{hu2017learning,hudson2018compositional}. Other approaches have included using relational networks (RNs)~\cite{santoro2017simple}, early fusion~\cite{malinowski2018visual}, and conditional feature transformations~\cite{perez2018film}. In our experiments, we evaluate RN~\cite{santoro2017simple} and  MAC~\cite{hudson2018compositional}, which are explained in more detail in Sec.~\ref{sec:vqa-models}.

Although rare exceptions exist~\cite{hu2017learning}, most of these algorithms are evaluated only on natural or synthetic VQA datasets and not both. Furthermore, several algorithms that claim specific abilities are not tested on datasets  designed to test these abilities, \eg, QCG~\cite{norcliffe2018learning} claims better compositional performance, but it is not evaluated on CVQA~\cite{Agrawal2017CVQAAC}. Here, we evaluate multiple state-of-the-art algorithms on both natural and synthetic VQA datasets, and we propose a new algorithm that works well for both. 

\section{The \gls{RAMEN} VQA Model}

\begin{figure}[t!]
    \centering
    \includegraphics[width=0.48\textwidth]{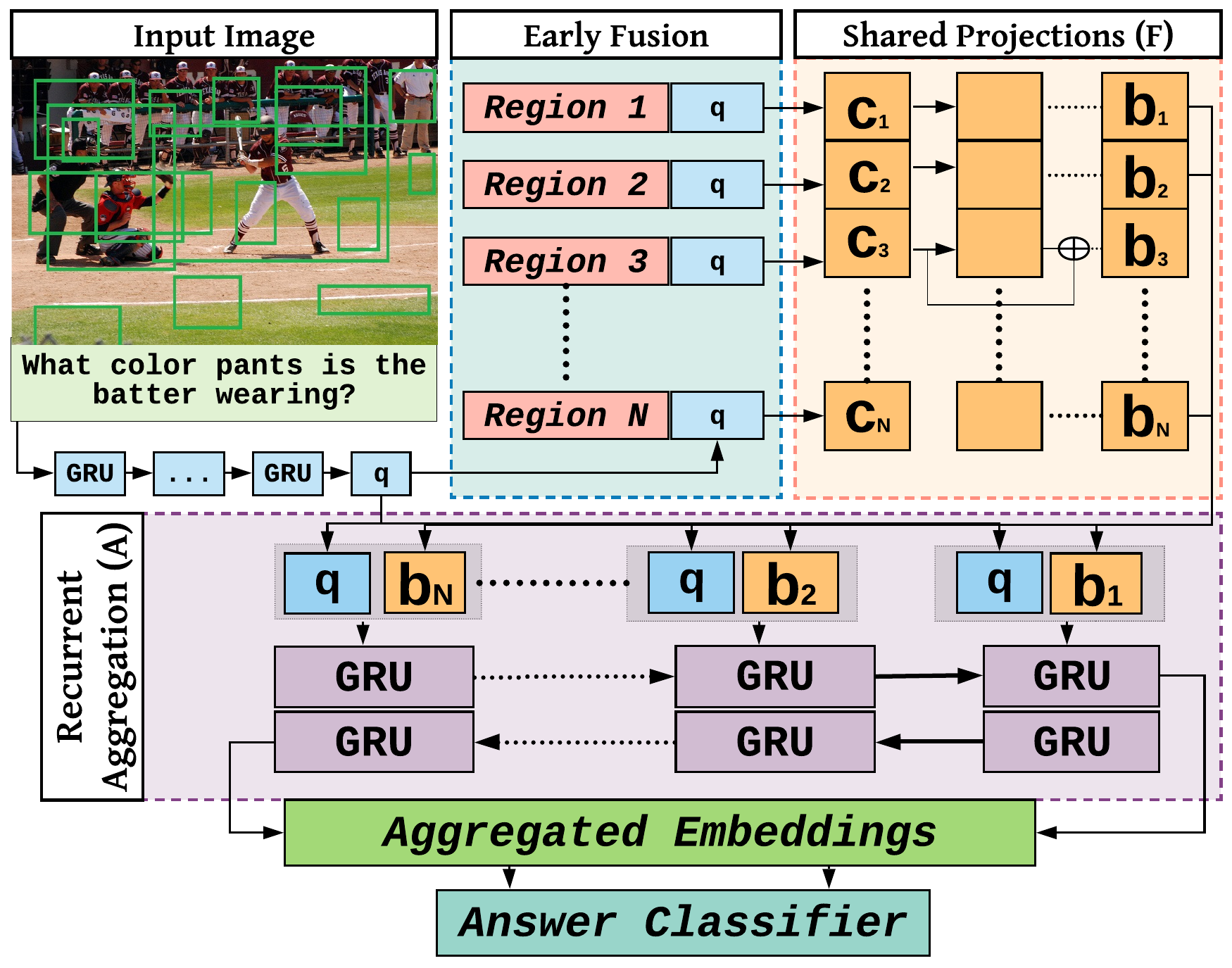}
    \caption{Our recurrent aggregation of multimodal embeddings network (RAMEN).}
    \label{fig:model-diagram}
\end{figure}

We propose the Recurrent Aggregation of Multimodal Embeddings Network (\gls{RAMEN}) model for VQA. It is designed as a conceptually simple architecture that can adapt to the complexity of natural scenes, while also being capable of answering questions requiring  complex chains of compositional reasoning, which occur in synthetic datasets like CLEVR.  As illustrated in Fig.~\ref{fig:model-diagram}, \gls{RAMEN} processes visual and question features in three phases:
\begin{enumerate}
    \item \textbf{Early fusion of vision and language features.} Early fusion between visual and language features and/or early modulation of visual features using language has been shown to help with compositional reasoning~\cite{malinowski2018visual,perez2018film,de2017modulating}. Inspired by these approaches, we propose early fusion through concatenation of spatially localized visual features with question features.
    \item \textbf{Learning bimodal embeddings via shared projections.} The concatenated visual+question features are passed through a shared network, producing spatially localized bimodal embeddings. This phase helps the network learn the inter-relationships between the visual and textual features. 
    \item \textbf{Recurrent aggregation of the learned bimodal embeddings.} We aggregate the bimodal embeddings across the scene using a bi-directional gated recurrent unit (bi-GRU) to capture interactions among the bimodal embeddings. The final forward and backward states essentially need to retain all of the information required to answer the question.
\end{enumerate}

While most recent state-of-the-art VQA models for natural images use attention~\cite{anderson2018bottom} or bilinear pooling mechanisms~\cite{kim2018bilinear}, RAMEN is able to perform comparably without these mechanisms.  Likewise, in contrast to the state-of-the-art models for CLEVR, RAMEN does not use pre-defined modules~\cite{Mascharka_2018_CVPR} or reasoning cells~\cite{hudson2018compositional}, yet our experiments demonstrate it is capable of compositional reasoning.

\subsection{Formal Model Definition} 

The input to RAMEN is a question embedding $\bm{q} \in \mathbb{R}^d$ and a set of $N$ region proposals $\bm{r}_i \in \mathbb{R}^m$, where each $\bm{r}_i$ has both visual appearance features and a spatial position.  RAMEN first concatenates each proposal  with question vector, which is followed by batch normalization, \ie,
\begin{equation}
\bm{c}_i = \textit{BatchNorm}\left( \bm{r}_i \oplus \bm{q} \right), 
\end{equation}
where $\oplus$ represents concatenation.

All $N$ of the $\bm{c}_i$ vectors are then passed through a function $F\left( \bm{c}_i \right)$, which mixes the features to produce a bimodal embedding $\bm{b}_i = F\left( \bm{c}_i \right)$, where  $F\left( \bm{c}_i \right)$ was modeled using a multi-layer perceptron (MLP) with residual connections. 

Next, we perform late-fusion by concatenating each bimodal embedding with the original question embedding and aggregate the collection using
\begin{equation}
\bm{a} = A\left( \bm{b}_1 \oplus \bm{q}, \bm{b}_2 \oplus \bm{q}, \ldots, \bm{b}_N \oplus \bm{q}  \right),
\end{equation}
where the function $A$ is modeled using a bi-GRU, with the output of $A$ consisting of the concatenation of the final states of both the forward and backward GRUs. We refer to $\bm{a}$ as the RAMEN embedding, which is then sent to a classification layer that predicts the answer. While RAMEN is simpler than most recent VQA models, we show it is competitive across datasets, unlike more complex models.

\subsection{Implementation Details}
\label{sec:implementation-details}
\paragraph{Input Representation.} We represent question words as 300 dimensional embeddings initialized with pre-trained GloVe vectors~\cite{pennington2014glove}, and process them with a GRU to obtain a 1024 dimensional question embedding, \ie, $\bm{q} \in \mathbb{R}^{1024}$. Each region proposal $\bm{r}_i \in \mathbb{R}^{2560}$ is made of visual features concatenated with spatial information. The visual features are 2048 dimensional CNN features produced by the bottom-up architecture~\cite{anderson2018bottom} based on Faster R-CNN~\cite{ren2015faster}. Spatial information is encoded by dividing each proposal into a $16 \times 16$ grid of $(x, y)$-coordinates, which is then flattened to form a 512-dimensional vector.  

\paragraph{Model Configuration.} The projector $F$ is modeled as a 4-layer MLP with 1024 units with swish non-linear activation functions~\cite{DBLP:journals/corr/abs-1710-05941}. It has residual connections in layers 2, 3 and 4. The aggregator $A$ is a single-layer bi-GRU that has a 1024 dimensional hidden state, so the concatenation of forward and backward states produces a 2048 dimensional embedding. This embedding is projected through a 2048 dimensional fully connected swish layer, followed by an output classification layer that has one unit per possible answer in the dataset. 

\paragraph{Training Details.}  RAMEN is trained with Adamax~\cite{Kingma2014AdamAM}. Following~\cite{kim2018bilinear}, we use a gradual learning rate warm up  ($2.5 * epoch * 10^{-4}$ ) for the first $4$ epochs,  $5 * 10^{-4}$ for epochs 5 to 10, and then decay it at the rate of $0.25$ for every $2$ epochs, with early stopping used.  The mini-batch size is 64.

\begin{figure*}[t!]
    \centering
    \includegraphics[width=0.95\textwidth]{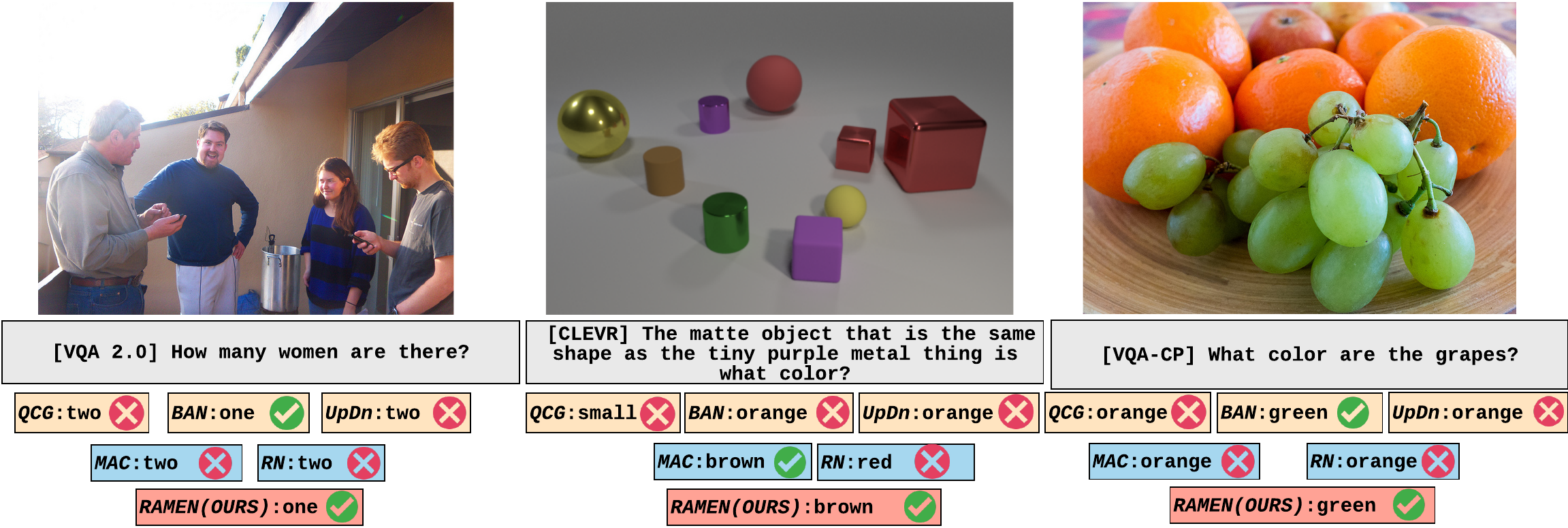}
    \caption{Some example predictions from our model RAMEN compared to other existing methods.}
    \label{fig:examples}
\end{figure*}

\section{VQA Models Evaluated} \label{sec:vqa-models}
In this section, we will briefly describe the models evaluated in our experiments.

\textbf{Bottom-Up-Attention and Top-Down (\gls{UpDn})}~\cite{anderson2018bottom} combines bottom-up and top-down attention mechanisms to perform VQA, with the bottom-up mechanism generating object proposals from Faster R-CNN~\cite{ren2015faster}, and the top-down mechanism predicting an attention distribution over those proposals. The top-down attention is task-driven, using questions to predict attention weights over the image regions. This model obtained first place in the 2017 VQA Workshop Challenge. For fair comparison, we use its bottom-up region features for all other VQA models.

\textbf{Question-Conditioned Graph (\qcg)}~\cite{norcliffe2018learning} represents images as graphs where object-level features from bottom-up region proposals~\cite{anderson2018bottom} act as graph nodes and edges that encode interactions between regions that are conditioned on the question. For each node, QC-Graph chooses a neighborhood of nodes with the strongest edge connections, resulting in a question specific graph structure. This structure is processed by a patch operator to perform spatial graph convolution~\cite{kipf2017graph}. The main motivation behind choosing this model was to examine the efficacy of the proposed graph representations and operations for compositional reasoning.

\textbf{Bilinear Attention Network (BAN)}~\cite{kim2018bilinear} fuses visual and textual modalities by considering interactions between all region proposals (visual channels) with all question words (textual channels). Unlike dual-attention mechanisms~\cite{nam2016dual}, BAN handles interactions between all channels. It can be considered a generalization of low-rank bilinear pooling methods that jointly represent each channel pair~\cite{LuYBP16,kim2016hadamard}. BAN supports multiple glimpses of attention via connected residual connections. It achieves 70.35\% on the test-std split of \gls{VQAv2}, which is one of the best published results.

\textbf{Relation Network (\gls{RN})}~\cite{santoro2017simple} takes in every pair of region proposals, embeds them, and sums up all $N^2$ pair embeddings to produce a vector that encodes relationships between objects. This pairwise feature aggregation mechanism enables compositional reasoning, as demonstrated by its performance on CLEVR dataset. However, RN's computational complexity increases quadratically with the number of objects, making it expensive to run when the number of objects is large. There have been recent attempts at reducing the number of pairwise comparisons by reducing the number of input objects fed to RN~\cite{malinowski2018learning,acharya2019tallyqa}.

The \textbf{Memory, Attention and Composition (\gls{MAC})} network~\cite{hudson2018compositional} uses computational cells that automatically learn to perform attention-based reasoning. Unlike,  modular networks~\cite{AndreasRDK15,hu2017learning,andreas2016learning} that require pre-defined modules to perform pre-specified reasoning functions, MAC learns reasoning mechanisms directly from the data. Each MAC cell maintains a control state representing the reasoning operation and a memory state that is the result of the reasoning operation. It has a computer-like architecture with read, write and control units. MAC was evaluated on the CLEVR datasets and reports significant improvements on the challenging counting and numerical comparison tasks.

\subsection{Standardizing Models}
Often VQA models achieve state-of-the-art performance using visual features that differ from past models, making it difficult to tell if good performance came from model improvements or improvements to the visual feature representation. To make the comparison across models more meaningful, we use the same visual features for all algorithms across all datasets. Specifically, we use the 2048-dimensional `bottom-up' CNN features produced by the region proposal generator of a trained Faster R-CNN model~\cite{girshick2015fast} with a ResNet-101 backend. Following~\cite{teney2018tips}, we keep the number of proposals fixed at 36 for natural images, although performance can increase when additional proposals are used, e.g., others have reported that using 100 proposals with BAN can slightly increase its performance~\cite{kim2018bilinear}. This Faster R-CNN model is trained for object localization, attribute recognition, and bounding box regression on Visual Genome~\cite{krishnavisualgenome}. While CNN feature maps have been common for CLEVR, state-of-the-art methods for CLEVR have also been shifting toward region proposals~\cite{nsvqa}. For datasets that use CLEVR's images, we train a separate Faster R-CNN for multi-class classification and bounding box regression, because the Faster R-CNN trained on Visual Genome did not transfer well to CLEVR. To do this, we estimate the bounding boxes using 3D coordinates/rotations specified in the scene annotations. We keep the number of CLEVR regions fixed at 15.  We also augment these features with a 512 dimensional vector representing positional information about the boxes as described in Sec.~\ref{sec:implementation-details} for \gls{TDIUC}, \gls{CLEVR}, \gls{CLEVR-Humans} and \gls{CoGenT}. Following~\cite{anderson2018bottom}, we limit the set of candidate answers to those occurring at least 9 times in the training+validation set, resulting in vocabularies of 2185 answers for \gls{VQAv1} and 3129 answers for \gls{VQAv2}. Following \cite{vqacp,Agrawal2017CVQAAC}, we limit the answer vocabulary to the 1000 most frequent training set answers for \gls{CVQA} and \gls{VQACP}. For \gls{VQAv2}, we train the models on training and validation splits and report results on test-dev split. For the remaining datasets, we train the models on their training splits and report performance on validation splits.



\begin{table*}[t]
\centering
\small
\caption{Overall results from six VQA models evaluated using same visual features across all datasets. We highlight the top-3 models for each dataset, using darker colors for better performers. To study the generalization gap, we present the results before fine-tuning  for CLEVR-CoGenT and CLEVR-Humans. For \gls{VQAv2}, we train models on the train and validation splits and report results on test-dev questions. For CLEVR-CoGenT-B, we report results on a sub-split of validation split. For the other  datasets, we train models on the train split and report results on validation splits.} \label{tbl:overall-results}
\begin{tabular}{@{}lcccccc@{}}
\toprule
\textbf{Dataset/Algorithm} & \textbf{UpDn}                 & \textbf{QCG} & \textbf{BAN}                                         & \textbf{MAC}                                     & \textbf{RN}                        & \textbf{Ours}                      \\ \midrule
\textbf{\gls{VQAv1}}              & \cellcolor[HTML]{EBECFF}  {60.62} & 59.90        & \cellcolor[HTML]{B1B2FF}\textbf{62.98}               & 54.08                                            & 51.84                              & \cellcolor[HTML]{C4C7FD} {61.98}         \\
\textbf{\gls{VQAv2}}              & \cellcolor[HTML]{EBECFF}64.55 & 57.08        & \cellcolor[HTML]{B1B2FF}\textbf{67.39}               & 54.35                                            & 60.96                              & \cellcolor[HTML]{C4C7FD} {65.96}          \\
\textbf{\gls{TDIUC}}             & \cellcolor[HTML]{EBECFF}68.82 & 65.57        & \cellcolor[HTML]{C4C7FD}71.10               & 66.43                                            & 65.06                              & \cellcolor[HTML]{B1B2FF} \textbf{{72.52}}          \\
\textbf{\gls{CVQA}}             & \cellcolor[HTML]{EBECFF}57.01 & 56.45        & \cellcolor[HTML]{C4C7FD}57.36               & 50.99                                            & 48.11                              & \cellcolor[HTML]{B1B2FF}\textbf{{58.92}}          \\
\textbf{\gls{VQACP}}            & 38.01 & \cellcolor[HTML]{EBECFF}38.32        & \cellcolor[HTML]{B1B2FF}\textbf{39.31}               & 31.96                                            & 26.70                              &  \cellcolor[HTML]{C4C7FD}{39.21}          \\ \midrule
\textbf{\gls{CLEVR}}             & 80.04                         & 46.73        & 90.79                                                & \cellcolor[HTML]{B1B2FF}\textbf{98.00}           & \cellcolor[HTML]{EBECFF}95.97      & \cellcolor[HTML]{C4C7FD} {96.92}          \\
\textbf{CLEVR-Humans}      & 54.51                         & 28.12        & \cellcolor[HTML]{B1B2FF}\textbf{60.23} & {\color[HTML]{000000}}50.20           & \cellcolor[HTML]{EBECFF}57.65      & \cellcolor[HTML]{C4C7FD}57.87          \\
\textbf{CLEVR-CoGenT-A}    & 82.47                             & 59.63        & 92.50                                                & \cellcolor[HTML]{B1B2FF}\textbf{98.04}           & \cellcolor[HTML]{EBECFF}96.45      & \cellcolor[HTML]{C4C7FD}{96.74}          \\
\textbf{CLEVR-CoGenT-B}    & 72.22 & 53.45        & 79.48                                                & \cellcolor[HTML]{B1B2FF}\textbf{90.41}           & \cellcolor[HTML]{EBECFF}84.68      & \cellcolor[HTML]{C4C7FD} {89.07}          \\  \bottomrule
\textbf{Mean}              & 64.18                  &  51.69           & \cellcolor[HTML]{C4C7FD}69.00                                                    &  \cellcolor[HTML]{EBECFF}{\color[HTML]{000000}66.05} & 65.26 & \cellcolor[HTML]{B1B2FF}\textbf{71.02} \\ \bottomrule
\end{tabular}
\end{table*}

\textbf{Maintaining Compatibility.} UpDn, QCG and BAN are all designed to operate on region proposals. For both MAC and RN, we needed to modify the input layers to accept bottom-up features, instead of convolutional feature maps. This was done so that the  same features could be used across all datasets and also to upgrade RN and MAC so that they would be competitive on natural image datasets where these features are typically used~\cite{anderson2018bottom}. For MAC, we replace the initial 2D convolution operation with a linear projection of the bottom-up features. These are fed through MAC's read unit, which is left unmodified. For RN, we remove the initial convolutional network and directly concatenate bottom-up features with question embeddings as the input. The performance of both models after these changes are comparable to the versions using learned convolutional feature maps as input, with MAC achieving 98\% and RN achieving 95.97\% on the CLEVR validation set.

\section{Experiments and Results}

\subsection{Main Results}

In this section, we demonstrate the inability of current VQA algorithms to generalize across natural and synthetic datasets, and show that \gls{RAMEN} rivals the best performing models on all datasets. We also present a comparative analysis of bias-resistance, compositionality, and generalization abilities for all six algorithms. Table~\ref{tbl:overall-results} provides our main results for all six algorithms on all eight datasets. We use the standard metrics for all datasets, \ie, we use simple accuracy for the CLEVR family of datasets, mean-per-type for TDIUC, and `10-choose-3' for VQAv1, VQAv2, CVQA, and VQACPv2. Some example outputs for RAMEN compared to other models are given in Fig.~\ref{fig:examples}.

\paragraph{Generalization Across VQA Datasets.}
RAMEN achieves the highest results on \gls{TDIUC} and \gls{CVQA} and is the second best model for VQAv1, VQAv2, VQACPv2 and all of the CLEVR datasets. On average, it has the highest score across datasets, showcasing that it can generalize across natural datasets and synthetic datasets that test reasoning. BAN achieves the next highest mean score. \gls{BAN} works well for natural image datasets, outperforming other models on \gls{VQAv1}, \gls{VQAv2} and \gls{VQACP}. However, \gls{BAN} shows limited compositional reasoning ability.  Despite being conceptually much simpler than \gls{BAN}, RAMEN outperforms \gls{BAN} by $6\%$ (absolute) on \gls{CLEVR} and  $10\%$ on CLEVR-CoGenT-B. \gls{RAMEN} is within 1.4\% of MAC on all compositional reasoning tests. \gls{UpDn} and \gls{QCG} perform poorly on \gls{CLEVR}, with \gls{QCG} obtaining a score below 50\%. 

\begin{table*}[t]
\caption{Performance comparison on TDIUC using three different metrics. MPT measures task generalization and N-MPT measures generalization to rare answers. We highlight the top-3 models, emboldening the winner.}
\centering
\small
\begin{tabular}{lrrrrrr}
\toprule
\textbf{Metric / Algorithm} & \textbf{UpDn} & \textbf{QCG} & \textbf{BAN} & \textbf{MAC} & \textbf{RN} & \textbf{Ours} \\ \midrule
\textbf{MPT} & \cellcolor[HTML]{EBECFF}68.82 & 65.67 & \cellcolor[HTML]{C4C7FD}71.10 & 66.43 & 65.06 & \cellcolor[HTML]{B1B2FF}\textbf{72.52} \\
\textbf{N-MPT} & 38.93 & 37.43 & \cellcolor[HTML]{C4C7FD}40.65 & \cellcolor[HTML]{EBECFF}39.02 & 35.75 & \cellcolor[HTML]{B1B2FF}\textbf{46.52} \\
\textbf{Simple Accuracy} & \cellcolor[HTML]{EBECFF}82.91 & 82.05 & \cellcolor[HTML]{C4C7FD}84.81 & 82.53 & 84.61 & \cellcolor[HTML]{B1B2FF}\textbf{86.86} \\ \bottomrule
\end{tabular}
\label{tbl:tdiuc}
\end{table*}

\begin{table*}[t]
\centering
\small
\caption{Performance on CLEVR's query types.}
\begin{tabular}{lccccccc}
\toprule
  & \multicolumn{1}{l}{\textbf{Exist}} & \textbf{\begin{tabular}[c]{@{}l@{}}Query \\ Attribute\end{tabular}} & \textbf{\begin{tabular}[c]{@{}l@{}}Compare \\ Attribute\end{tabular}} & \textbf{\begin{tabular}[c]{@{}l@{}}Equal \\ Integer\end{tabular}} & \textbf{\begin{tabular}[c]{@{}l@{}}Greater \\ Than\end{tabular}} & \textbf{\begin{tabular}[c]{@{}l@{}}Less \\ Than\end{tabular}} & \multicolumn{1}{l}{\textbf{Count}} \\ 

 \midrule
\textbf{UpDn} & 83.07 & 90.08 & 79.87 & 65.65 & 80.43 & 85.76 & 64.03 \\ 
\textbf{QCG} & 66.11 & 31.11 & 51.47 & 59.76 & 69.35 & 70.57 & 44.19 \\ 
\textbf{BAN} & 94.72 & 90.56 & 98.44 & 72.35 & 81.35 & 86.39 & 86.47 \\ 
\textbf{MAC} & 99.18 & 99.59 & 99.33 & 85.44 & 96.82 & 97.55 & 95.46 \\ 
\textbf{RN} & 98.40 & 98.19 & 97.81 & 77.30 & 93.40 & 84.27 & 90.90 \\ 
\textbf{RAMEN} & 98.90 & 98.93 & 99.30 & 79.40 & 93.41 & 88.53 & 94.10 \\ \bottomrule
\end{tabular}
\label{tbl:clevr}
\end{table*}

\paragraph{Generalization Across Question Types.} We use TDIUC to study generalization across question types. TDIUC has multiple accuracy metrics, with mean-per-type (MPT) and normalized mean-per-type (N-MPT) compensating for biases.  As shown in Table~\ref{tbl:tdiuc}, all methods achieve simple accuracy scores of over 82\%; however, both MPT and N-MPT scores are 13-20\% lower. Lower MPT scores indicate that all algorithms are struggling to generalize to multiple tasks. \gls{RAMEN} obtains the highest MPT score of 72.52\% followed by \gls{BAN} at 71.10\%. For all algorithms, `object presence,' `object recognition,' and `scene recognition' are among the easiest tasks, with all of the methods achieving over 84\%  accuracy on them; however, these tasks all have relatively large amounts of training data (60K - 657K QA pairs each). All of the methods performed well on `sports recognition' (31K QA pairs), achieving over 93\%, but all performed poorly on a  conceptually similar task of `activity recognition' (8.5K QA pairs), achieving under 62\% accuracy. This showcases the inability  to generalize to question types with fewer examples. To emphasize this, TDIUC provides the Normalized MPT (N-MPT) metric that measures generalization to rare answers by taking answer frequency into account. The differences between normalized and un-normalized scores are large for all models. \gls{RAMEN} has the smallest gap, indicating a better resistance to answer distribution biases, while \gls{BAN} has the largest gap.

 \paragraph{Generalization to Novel Concept Compositions.}
 We evaluate concept compositionality using CVQA and CLEVR-CoGenT-B.  As shown in \tbl~\ref{tbl:overall-results},  scores on CVQA are lower than VQAv1, suggesting all of the algorithms struggle when combining concepts in new ways.  \gls{MAC} has the largest performance drop, which suggests  its reasoning cells were not able to compose real-world visuo-linguistic concepts effectively.

To evaluate the ability to generalize to new concept compositions on the synthetic datasets, we train the models on CLEVR-CoGenT-A's train split and evaluate on the validation set without fine-tuning. Following \cite{perez2018film}, we obtain a test split from the validation set of `B,' and report performance without fine-tuning on `B.' All algorithms show a large drop in performance. Unlike the CVQA results, MAC's drop in performance is smaller. Again, RAMEN has a comparatively small decrease in performance. 

\paragraph{Performance on \gls{VQACP}'s Changing Priors.}
All algorithms have a large drop in performance under changing priors. This suggests there is significantly more work to be done to make VQA algorithms overcome linguistic and visual priors so that they can more effectively learn to use generalizable concepts.

\paragraph{Counting and Numerical Comparisons.}
For \gls{CLEVR}, counting and number comparison (`equal integer,' `greater than,' and `less than') are the most challenging tasks across algorithms as shown in \tbl~\ref{tbl:clevr}. \gls{MAC} performs best on these tasks, followed by \gls{RAMEN}. Algorithms apart from \gls{MAC} and \gls{QCG} demonstrate a large  ($> 4.8\%$) discrepancy between `less than' and `greater than' question types, which require similar kinds of reasoning. This discrepancy is most pronounced for \gls{RN} (9.13\%), indicating a difficulty in linguistic understanding. \gls{BAN} uses a counting module~\cite{zhang2018vqacount}; however, its performance on CLEVR's counting task is still 9\% below \gls{MAC}. All of the algorithms struggle with counting in natural images too. Despite TDIUC having over 164K counting questions, all methods achieve a score of under 62\% on these questions. 

\paragraph{Other CLEVR Tasks.}
As shown in \tbl~\ref{tbl:clevr}, \gls{RAMEN} is within 0.03-1.5\% of \gls{MAC}'s performance on all tasks except number comparison. \gls{UpDn} and \gls{QCG} are the worst performing models on all query types.  Except for \gls{QCG}, all of the models find it easy to answer queries about object attributes and existence. Models apart from \gls{UpDn} and \gls{QCG} perform well on attribute comparison questions that require comparing these properties. Surprisingly, \gls{BAN} finds attribute comparison, which requires more reasoning, easier than the simpler attribute query task.
We present results on CLEVR-Humans without fine-tuning to examine how well algorithms handle free-form language if they were only trained on CLEVR's vocabulary. \gls{BAN} shows the best generalization, followed by \gls{RAMEN} and \gls{RN}.

\begin{table}[t]
\centering
\small
\caption{Ablation studies comparing early versus late fusion between visual and question features, and comparing alternate aggregation strategies. \label{tbl:ablation-study}}

\begin{tabular}{lcc}
\toprule
 & \textbf{VQAv2} & \textbf{CLEVR} \\ 
 \midrule
\textbf{Without Early Fusion} & 61.81  & 77.48  \\ 
\textbf{Without Late Fusion} & 65.64 & 96.63 \\ 
\textbf{Aggregation via Mean Pooling} & 63.01 & 92.45 \\
\midrule
\textbf{Without Ablation} & 65.96 & 96.92 \\
\bottomrule
\end{tabular}
\end{table}
\subsection{Ablation Studies}
Results from several ablation studies to test the contributions of RAMEN's components are given in \tbl~\ref{tbl:ablation-study}. We found that early fusion is critical to RAMEN's performance, and removing it causes an almost 20\% absolute drop in accuracy for CLEVR and a 4\% drop for \gls{VQAv2}. Removing late fusion has little impact on CLEVR and \gls{VQAv2}. 

We also explored the utility of using a bi-GRU for aggregation compared to using mean pooling, and found that this caused a drop in performance for both datasets. We believe that the recurrent aggregation aids in capturing interactions between the bimodal embeddings, which is critical for reasoning tasks, and that it also helps remove duplicate proposals by performing a form of non-maximal suppression.

\subsection{Newer Models}

Additional VQA algorithms have been released since we began this project, and some have achieved higher scores than the models we evaluated on \emph{some} datasets. The Transparency By Design (TBD) network~\cite{Mascharka_2018_CVPR} obtains 99.10\%  accuracy on \gls{CLEVR} by using  ground truth functional programs to train the network, which are not available for natural VQA datasets. Neural-Symbolic VQA (NS-VQA)~\cite{nsvqa} reports a score of 99.80\% on CLEVR, but uses a question parser to allocate functional modules along with highly specialized segmentation-based CNN features. They did not perform ablation studies to determine the impact of using these visual features. None of the models we compare have access to these additional resources.

Results on VQAv2 can be significantly improved by using additional data from other VQA datasets and ensembling, \eg, the winner of the 2018 challenge used dialogues from Visual Dialog~\cite{das2017visual} as additional question answer pairs and an ensemble of 30 models. These augmentations could be applied to any of the models we evaluated to improve performance.  \gls{VQACP} results can also be improved using specialized architectures, \eg GVQA~\cite{vqacp} and \gls{UpDn} with adversarial regularization~\cite{DBLP:journals/corr/abs-1810-03649}. However, their performance on \gls{VQACP} is still poor, with \gls{UpDn} with adversarial regularization obtaining 42.04\% accuracy, showing only 2.98\% improvement over the non-regularized model.

\section{Discussion: One Model to Rule them All?}

We conducted the first systematic study to examine if the VQA systems that work on synthetic datasets generalized to real-world datasets, and vice versa. This was the original scope of our project, but we were alarmed when we discovered none of the methods worked well across datasets. This motivated us to create a new algorithm. Despite being simpler than many algorithms, \gls{RAMEN} rivals or even surpasses other methods. We believe some state-of-the-art architectures  are likely over-engineered to exploit the biases in the domain they were initially tested on, resulting in a deterioration of performance when tested on other datasets. This leads us to question whether the use of highly specialized mechanisms that achieve state-of-the-art results on one specific dataset will lead to significant advances in the field, since our conceptually simpler algorithm performs competitively across both natural and synthetic datasets without such mechanisms.

We advocate for the development of a single VQA model that does well across a wide range of challenges. Training this model in a continual learning paradigm would assess forward and backward transfer~\cite{hayes2019memory,kemker2018measuring,parisi2019continual}. Another interesting avenue is to combine VQA with related tasks like visual query detection~\cite{acharya2019vqd}. Regardless, existing algorithms, including ours, still have a long way to go toward showcasing both visuo-linguistic concept understanding \emph{and} reasoning. As evidenced by the large performance drops on \gls{CVQA} and \gls{VQACP}, current algorithms perform poorly at learning compositional concepts and are affected by biases in these datasets, suggesting reliance on superficial correlations. We observed  that methods developed solely for synthetic closed-world scenes are often unable to cope with unconstrained natural images and questions. Although performance on VQAv2 and CLEVR are approaching human levels on these benchmarks, our results show VQA is far from solved. We argue that future work should focus on creating one model that works well across domains. It would be interesting to train a dataset on a universal training set and then evaluate it on multiple test sets, with each test set demanding a different skill set. Doing so would help in seeking one VQA model that can rule them all.

\section{Conclusion}
Our work endeavors to set a new standard for what should be expected from a VQA algorithm: good performance across both natural scenes and challenging synthetic benchmarks. We hope that our work will lead to future advancements in VQA. 

\paragraph{Acknowledgements.} We thank NVIDIA for the GPU donation. This work was supported in part by a gift from Adobe Research.

{\small
\bibliographystyle{ieee}
\bibliography{egbib}
}
\end{document}